\documentclass[acmtist]{acmart}
\setcopyright{none}
\renewcommand\footnotetextcopyrightpermission{}
\usepackage{makecell}
\usepackage{caption}
\usepackage{subcaption}
\usepackage{algpseudocode}
\usepackage[section]{algorithm}

\AtBeginDocument{%
  }

\title{Federated Gaussian Mixture Models}
\author{Sophia Zhang Pettersson}
\affiliation{
  \institution{Cloud and Embedded Platform, Traton AB, Södertälje}
  \city{Södertälje}
  \country{Sweden}
}
\email{sophia.zhang.pettersson@scania.com}

\author{Kuo-Yun Liang}
\affiliation{
  \institution{Cloud and Embedded Platform, Traton AB, Södertälje}
  \city{Södertälje}
  \country{Sweden}
}
\email{kuo-yun.liang@scania.com}

\author{Juan Carlos Andresen}
\affiliation{
  \institution{Cloud and Embedded Platform, Traton AB, Södertälje}
  \city{Södertälje}
  \country{Sweden}
}
\email{juan-carlos.andresen@scania.com}

\begin{document}

\begin{abstract}
This paper introduces FedGenGMM, a novel one-shot federated learning approach for Gaussian Mixture Models (GMM) tailored for unsupervised learning scenarios. In federated learning (FL), where multiple decentralized clients collaboratively train models without sharing raw data, significant challenges include statistical heterogeneity, high communication costs, and privacy concerns. FedGenGMM addresses these issues by allowing local GMM models, trained independently on client devices, to be aggregated through a single communication round. This approach leverages the generative property of GMMs, enabling the creation of a synthetic dataset on the server side to train a global model efficiently. Evaluation across diverse datasets covering image, tabular, and time series data demonstrates that FedGenGMM consistently achieves performance comparable to non-federated and iterative federated methods, even under significant data heterogeneity. Additionally, FedGenGMM significantly reduces communication overhead, maintains robust performance in anomaly detection tasks, and offers flexibility in local model complexities, making it particularly suitable for edge computing environments.

\end{abstract}


\maketitle

\section{Introduction}
Federated learning (FL) is a concept in machine learning (ML) that allows multiple devices or clients to collaboratively train a shared model, without having to share their data with each other or with a central server \cite{mcmahan}. Clients keep their data locally and only communicate model parameters or updates. FL can thus be useful in scenarios where it is impractical or even impossible to transfer large amounts of data to a central location, for instance, due to communication costs or for reasons of privacy and security. It is common to make a distinction between cross-silo FL (that is, cooperation between relatively few clients, such as hospitals with large computational capabilities) and cross-device FL, where the number of clients (such as mobile devices or connected vehicles) is typically large and each client has limited resources for communication and computations, with potentially unreliable connectivity \cite{Kairouz_2021}. FL scenarios can also be characterized by how the data is partitioned among the clients. In vertical partitioning, clients share samples but have different sets of features. In horizontal partitioning, the set of features is the same between all clients, but the samples are different \cite{Kairouz_2021}. In this paper, we are mainly concerned with the cross-device, horizontal scenario, where the clients are assumed to be edge devices of some kind.

FL has great potential, but the successful realization of the concept also faces specific challenges. An important and much-studied issue in this context is how to handle the statistical heterogeneity of the data distribution between clients \cite{Sattler2019ClusteredFL}. Especially in the context of cross-device FL, clients typically come in many different flavors and are operated in different environments by different users. Other FL challenges come in the form of communication delays, high communication costs, and security issues such as information leakage and adversarial attacks \cite{sattler_fl, Wei_2020}.

Most FL algorithms are iterative, comprising several rounds of communication for the exchange of model parameters or updates. In contrast, the one-shot FL approach \cite{guha_one-shot_fl} has been proposed as a way of alleviating both communication and security issues by limiting communication between parties to a single round.

Most existing studies on FL have focused on supervised learning, whereas unsupervised learning and other categories of ML have received less attention in this context \cite{Kairouz_2021, non_IID_CL_2022, dong_fadngs_fl_ad}. Furthermore, the typical FL setup in the literature employs some form of neural network (NN) model rather than statistical ML models \cite{fed_tab, pandhare}, and the vast majority of FL use cases revolve around image classification, using datasets such as MNIST, FEMNIST, CIFAR, et al. \cite{non_IID_CL_2022}. Thus, there are a substantial number of studies (e.g., \cite{FedProx_Li_2020, Marfoq_2021, modfl, pfl_mixture, vucovich_fl_ad}) on how to extend and improve the FedAvg algorithm, originally presented in \cite{mcmahan}, in this context. However, FL approaches for statistical ML models appear to be less explored, especially in the areas of unsupervised learning and tabular data. An example of a type of ML model with proven strengths in these two areas is the Gaussian Mixture Model (GMM) \cite{figueiredo_2002_mixture_models, kasa_misspecified_gmms_2023}.

In this paper, we apply FedGenGMM, a one-shot approach, to the problem of federated GMM learning. Specifically, models are trained locally until convergence, and then the model parameters are sent to the server where we utilize the generative property of the GMM to perform model aggregation. The method is straightforward to implement, does not require any sharing of raw data, and uses less communication and synchronization resources than iterative FL methods during the training phase. We evaluate FedGenGMM by analyzing the ability of the shared model to capture the patterns in the local data and by using the model to perform anomaly detection on several datasets, both from the public domain and from the industry, under different conditions of heterogeneity. The topic of anomaly detection in an unsupervised federated setting is of interest in itself and has not been thoroughly studied so far. Furthermore, we evaluate the solutions using both image, tabular, and time-series data.

\section{Related work}

\subsection{Federated GMM}
The GMM belongs to the family of finite-mixture models. The standard method for training such models is known as the Expectation Maximization (EM) algorithm, first introduced in \cite{DEMP1977} and since then extended and improved in a large body of research, e.g. \cite{figueiredo_2002_mixture_models, split_and_merge_EM, INGRASSIA_EM_GMM_2011, kasa_misspecified_gmms_2023}.
In its original form, the EM algorithm assumes that all data is available at the location where the model training takes place. However, although relatively less explored than FL for NN models, federated EM algorithms for GMM training have been proposed in several previous studies, either as a central concept \cite{pandhare} or as a tool for improving FL of NN models under heterogeneous conditions \cite{pfl_mixture}. In \cite{pandhare}, the authors developed a distributed version of the EM algorithm (DEM) for the training of GMMs, with the aim of detecting anomalies in an industry setting. In \cite{ht-fed-gan}, a distributed algorithm for variational Bayesian GMM was deployed to facilitate the generation of synthetic data based on generative adversarial networks (GANs) for tabular datasets. The authors of \cite{pfl_mixture} utilized GMMs to enhance the FL of NN models for image classification under heterogeneous conditions. Here, the GMM training was an integral part of the NN model training, again using a DEM algorithm. In particular, they also pointed out the possibility of using the distributed GMM training concept independently.

A shared trait of the aforementioned solutions is thus that they are based on distributed versions of the EM algorithm, where each iteration involves sending parameter updates between clients and server. Furthermore, the GMM structure is assumed to be the same across all clients and the server, e.g., using the same number of GMM components and the same type of covariance matrix.

\subsection{One-shot FL}
In recent years, one-shot FL has emerged as an alternative to the iterative FL approach to supervised learning. In contrast to iterative methods, one-shot FL methods use only one single communication round to exchange information between clients and the server. This has potential benefits in the form of reduced communication cost and a lower risk of security breaches \cite{zhang_dense}. However, the approach also presents challenges in the presence of heterogeneity of clients \cite{heinbaugh2023datafree}.

In \cite{guha_one-shot_fl}, the authors used both model ensemble selection techniques and knowledge distillation to aggregate locally trained support vector machine (SVM) models (instead of simply averaging the parameters of the local models). However, heterogeneous data distributions were not considered. For the knowledge distillation part, a dataset known to all clients was used. For privacy reasons, this approach may be unfeasible in many cases \cite{zhu_data-free}. The method presented in \cite{Li2020PracticalOF} also employs public datasets to select ensemble models at the central level. An ensemble approach which does not require any public dataset is described in \cite{wang2023datafree}, where the client models are clustered at the server. Cluster representatives are selected based on criteria such as local validation accuracy and size of the local training dataset.

In the context of NN models and image classification, several studies have explored the technique of performing one-shot FL by sending information about local data distributions to the server in order to generate training data for a global model. An approach is illustrated in \cite{zhou_DOSFL}, where distilled versions of local data are sent to the server to be used for training, achieving large savings in communication costs. The authors of \cite{kasturi_fusion} instead let each client choose among a fixed set of typical distributions according to which one best fits their local dataset and then send the estimated distribution parameters (together with the local models) to generate training data on the server. A somewhat related design can be found in \cite{beitollahi_param_transfer}, a recent work in which GMMs are trained locally on class-conditional features generated by local models. The GMM parameters are then sent to the server for training of the global model. In \cite{zhang_dense}, only the trained models are uploaded to the server, where the global model is created in two steps. First, the uploaded client models are employed to train a generator with the aim of creating a synthetic dataset such that it has similar properties to the local datasets. In the next step, the synthetically generated images and the local models are used to train the final global model (using knowledge distillation). The approach in \cite{zhang_dense} allows for heterogeneous local model structures. A recent study that explicitly addresses the issue of strongly heterogeneous client distributions in one-shot FL for NN-based image classification is presented in \cite{heinbaugh2023datafree}, where conditional variational autoencoders are trained locally. The resulting weights of the decoding network are sent to the server, where they are used to create a global decoder. The global decoder is then used to create a synthetic dataset that is used to train the final global model.

The work of \cite{Dennis2021HeterogeneityFT} is one of the few that has considered one-shot FL in the context of unsupervised learning. Here, the authors develop a method for federated clustering in which the clients share their locally computed cluster centers with the server. The server then applies a clustering algorithm to the local centers, which produces the global centers. The method assumes that the centers are well separated in some sense. Under this condition, higher levels of heterogeneity is found to actually improve the performance of the clustering algorithm.


\subsection{Unsupervised FL and anomaly detection}
The topic of unsupervised FL methods is less well studied than FL in a classification setting \cite{nardi_unsupervised_fl_ad, dong_fadngs_fl_ad}. However, a study closely related to our work is \cite{pandhare} in which the authors present an unsupervised FL approach to the problem of anomaly detection, using a federated version of the EM algorithm for GMM training. Their overall approach is thus similar to ours but uses an iterative FL method, as opposed to our one-shot approach.

In the context of NN-based models, the work in \cite{vucovich_fl_ad} describes an unsupervised FL method for intrusion detection based on autoencoders. Here, heterogeneity in client distributions is handled by a federated feature scaling scheme, and the training process is modified to account for differences in local data set sizes. In \cite{nardi_unsupervised_fl_ad}, the authors devised a two-step method for unsupervised FL under heterogeneous conditions. First, locally trained one-class SVM models were used to cluster clients that have the same inlier distributions. Secondly, FedAvg was used for training a federated autoencoder, only averaging parameters of clients in the same cluster. One downside of their method is that it requires the exchange of SVM model parameters between every pair of clients. A recent example of unsupervised FL for anomaly detection under heterogeneous conditions can be found in \cite{dong_fadngs_fl_ad}, where each client makes a rough estimate of its local distribution of features in the form of a Gaussian mixture. The mixture parameters are then sent to the server, where they are aggregated and used to improve the performance of the anomaly detection in each client during the federated learning process. Finally, the global NN model is aggregated using a modified version of FedAvg. In the experimental evaluation in \cite{dong_fadngs_fl_ad}, the Gaussian mixtures were limited to having only one component.

In addition, some works (e.g. \cite{berlo_unsupervised_fl_repr, Zhang2020FederatedUR}) have also considered unsupervised federated representation learning, where the purpose of the learning process is to extract data features that can be used for downstream tasks.

In this work, we consider the problem of using unsupervised FL to train GMMs with varying numbers of components. Training is performed using a communication-efficient one-shot approach, in a setting with heterogeneous client data distributions. 



\section{Problem formulation}
Suppose we have a set of $N$ clients $C$, where each client $c \in C$ has access to a local dataset $D_c$. Each dataset $D_c$ consists of $d$-dimensional data points $\{\textbf{x}_i \in \mathcal{R}^d, i=1,2, ..., |D_c|\}$, drawn from the distribution $p_c(\textbf{x})$. Let $D = \cup_{c \in C} D_c$. Our goal is to obtain a single centrally located GMM $G$ that models the distribution of $D$ (denoted as $p_D$), without sending any data points from the clients. $G$ may then be distributed to the clients, where the representations learned by the global model can be used for downstream tasks such as anomaly detection and classification. One example of a situation where this might be useful (instead of just relying on locally trained models) is if during training each client only operates in a limited number of environments, but during inference most of the clients can expect to be exposed to a significantly wider range of environments. Furthermore, new clients who join the scheme after completion of the model training could likely benefit from having access to an already trained global model.

The distributions of the local datasets $D_c$ can be heterogeneous among clients, and the heterogeneity is assumed to be in the form of feature distribution skew \cite{Kairouz_2021}. As is common in the FL literature (cf. \cite{Marfoq_2021, pfl_mixture, Li2022_non_iid_silos}), we further assume that each local distribution $p_c$ is a mixture of $M$ underlying, unknown distributions $p^{(m)}$, that is,

\begin{equation}
    \label{eq:input_mixture}
    p_c(\textbf{x}) = \sum_{m=1}^M \pi_{c,m} \cdot p^{(m)}(\textbf{x})\;,
\end{equation} where $\pi_{c,m}$ is the mixing weight of the underlying distribution $p^{(m)}$ for client $c$.

In the setting of density estimation, our aim is to have the $G$ model $p_D$ as closely as possible, given some specific configuration of G (such as the number of components and the type of covariance matrix). In this context, we will use the average log-likelihood to measure how a trained GMM fits the target distribution, that is, for a dataset $D_T = \{\textbf{x}_i, i = 1,2,...,T\}$, drawn from $p_D$, the problem can be described as maximizing the fitness score $\gamma_G$:

\begin{equation}
    \gamma_G(D_T) = \frac{1}{T} \sum_{i=1}^T f_G(\textbf{x}_i)\;,
\end{equation} where $f_G(\cdot)$ is the log-likelihood for one data point, as given by the global GMM $G$. That is, a higher score means that the model fits the data more closely (cf. \cite{DPGMM_2016}). In line with other works on one-shot FL (cf. \cite{guha_one-shot_fl, wang2023datafree, heinbaugh2023datafree}), during evaluation of the global model, the input samples will be drawn from the global distribution $p_D$.

In the setting of anomaly detection, we suppose that the input data can come from two distinct distributions with $p_D$ representing the "in" distribution (corresponding to non-anomalous data) and $p_{OOD}$ the "out" distribution (corresponding to anomalous data). We assume that clients only encounter (heterogeneous) data drawn from $p_D$ during training and creation of $G$. The global model is then applied to data drawn from both $p_D$ and $p_{OOD}$, our aim being to achieve a high degree of separation between the fitness scores for non-anomalous (inlier) data and the scores for anomalous (out-of-distribution, OOD) data, respectively.

\section{FedGenGMM}

\subsection{Method description}
We utilize the standard EM algorithm \cite{DEMP1977, Murphy2012MachineL, pfl_mixture} to train one local GMM in each client. After training, each local model $G_c$ will have the density

\begin{equation}
    p_{G_c}(\textbf{x}|\theta_c) = \sum_{k=1}^{K_c} r_{ck} \mathcal{N}(\textbf{x}|\mu_{ck}, \Sigma_{ck})\;,
\end{equation} where $\theta_c = \{r_{ck}, \mu_{ck}, \Sigma_{ck}\}, k = 1, 2, ..., K_c$, are the GMM parameters of $G_c$ with $K_c$ being the number of GMM components. Note that the number of components can be different for different clients. $r$, $\mu$, and $\Sigma$ represent the weight, the center, and the covariance matrix of each component in $G_c$, respectively.


When a client completes its local training (that is, when convergence has been reached), it sends the parameters of its local GMM $G_c$ to the server. After receiving the model parameters from the clients, the server uses the parameters to generate a synthetic dataset $S$. To promote similarity between the densities of $S$ and the global dataset $D$, the incoming mixture weights $r_{ck}$ for each client model are first modified according to the relative size of the corresponding local dataset $D_c$:

\begin{equation}
    \label{eq:r_ck}
    \forall c: r_{ck} \leftarrow r_{ck} \frac{|D_c|}{|D|}, k=1,2,...,K_c\;.
\end{equation} Note that this requires the server to be aware of the local dataset sizes and that it would of course be possible to use other types of client weighting schemes here. Then, one single GMM consisting of all the re-weighted components is created, and its weights are normalized. Finally, $S$ is created by drawing from the distribution given by the single GMM. The number of datapoints drawn is proportional to the number of incoming components:

\begin{equation}
    \label{eq:ns}
    |S| = H\sum_{c=1}^C K_c\;,
\end{equation} where the constant $H$ is a FedGenGMM hyperparameter, which thus together with the complexity of the local models determines the size of the synthetic dataset.

Finally, the EM algorithm is applied to train the global GMM $G$ on $S$. In this step, the number of components in $G$ and the type of covariance matrix can be fixed beforehand or determined during training, for example, by minimizing the Bayesian Information Criterion (BIC) score or using other strategies \cite{figueiredo_2002_mixture_models, split_and_merge_EM}. In the algorithm \ref{fedgengmm-algorithm} described below, this is exemplified by minimizing the BIC criterion both in the local training phase and in the aggregation phase. However, the main takeaway is that the hyperparameters of the local models and the global model can all be set independently of each other. Another benefit of FedGenGMM is that it would be fairly straightforward to replace the standard EM algorithm with another method to train local GMMs. A recent example of such a method is presented in \cite{kasa_misspecified_gmms_2023}, where the authors develop a method that shows resilience under conditions of noisy data and overlapping GMM clusters.


\subsection{FedGenGMM algorithm}
One example of a realization of FedGenGMM is shown in algorithm \ref{fedgengmm-algorithm}. Here, during the training of each client $c$, the BIC score is used to select the number of local GMM components ($K_c$) from a range of values (denoted by [$K_{\text{min}}, K_{\text{max}}$]) that is given as input to the algorithm. $K_c$ could be determined in other ways, without affecting the overall algorithm. The same observations apply to the training of the aggregated global model in the last step.

\begin{algorithm}[]
  \caption{FedGenGMM algorithm}
  \label{fedgengmm-algorithm}
  \begin{algorithmic}[1]
    \State \textbf{Input:} Set of clients ($C$) with local datasets $\{D_c\}$, Size of generated data ($H$)
    \State \textbf{Input:} Range of number of GMM components ($K_{\text{min}}, K_{\text{max}}$), Desired covariance type
    \State \textbf{Output:} Parameters of trained global model $G$

    \Procedure{TrainGMM}{$K_{\text{candidates}}$, dataset $X$}
        \State Let bic$_\text{min} = \infty$;
        \For{k in [$K_{\text{min}}, K_{\text{max}}$]} \Comment{Estimate $K$ using BIC}
            \State Let $Q$ be a GMM with desired covariance type and $K$ = k;
            \State Initialize the parameters of $Q$ using $X$;
            \State Use EM to train $Q$ on $X$;
            \If{BIC($Q$, $X$) < bic$_\text{min}$}
                \State Let bic$_\text{min}$ = BIC($Q$, $X$);
                \State Let $R = Q$;
            \EndIf
        \EndFor
        \State Return $R$;
    \EndProcedure
    
    \Procedure{FedGenGMM}{}
        \For{c = 1 to $|C|$} \Comment{Local training part for all clients}
            \State Let $G_c$ = TrainGMM($K_{\text{candidates}}$, $D_c$);
            \State Send parameters of $G_c$ and size of local dataset $|D_c|$ to the server;
        \EndFor

        \State Let $G_\text{tmp}$ = [an empty GMM]; \Comment{Server side aggregation part}
        \State Let $s = \sum_{i=1}^{|C|}|D_c|$;
        \For{c = 1 to $|C|$}
            \For{k = 1 to $K_c$}
                \State Add ($r_{ck} \frac{|D_c|}{s}$, $\mu_{ck}$, $\Sigma_{ck}$) to $G_\text{tmp}$;
            \EndFor
        \EndFor
        \State Normalize the component weights of $G_\text{tmp}$;
        \State Create $S$ by drawing $H \cdot \sum_{c=1}^C K_c$ values from $p_{G_\text{tmp}}$;
        \State Let $G$ = TrainGMM($K_{\text{candidates}}$, $S$);
        \State Return $G$;
    \EndProcedure
  \end{algorithmic}
\end{algorithm}

\subsection{Computational complexity and communication cost}
\label{sec:complexity}
The following analysis does not include the operations that may be needed to estimate $K$, neither in the clients nor in the server. The computational complexity in the clients is equal to the complexity of the standard EM-algorithm for training GMMs, that is $\mathcal{O}(|D_c| \cdot K_c \cdot \delta)$ per iteration \cite{Murphy2012MachineL}. Here $\delta$ is a term that depends on $d$ (the dimensionality of the dataset) and on the type of covariance matrix $\Sigma$. For full covariance, $\delta = d^2$, and for diagonal covariance $\delta = d$. (We note that the M-step typically involves inversion of the covariance matrix for each component of the GMM. This operation has complexity $\mathcal{O}(K_c \cdot d^3)$ if full covariance matrices are used. However, in the settings we consider, $d << |D_c|$, therefore, this term can be ignored.) 

The merging process in the server consists of two main steps. First, synthetic data is generated for each incoming GMM $G_c$. The generation of one such data point has complexity $\mathcal{O}(K_c + \delta)$, giving an overall complexity for the first step of $\mathcal{O}(|S| \cdot (K_c + \delta))$, where $|S|$ is the size of the generated dataset. Second, the standard EM algorithm is used to generate the global model $G$, which requires $\mathcal{O}(|S| \cdot K \cdot \delta)$ operations per EM iteration (where $K$ is the number of components in $G$). Thus, since $|S|$ increases linearly with the number of clients $C$ (Eq. \ref{eq:ns}), the server-side complexity will be linear with respect to $C$.

As for communication costs, the one-shot FedGenGMM process requires only one single round of communication to create the global model.

\subsection{Privacy}\label{sec:privacy}
The FedGenGMM process requires that clients share their local model parameters with the central server, causing statistical information about the local datasets to be exposed. In scenarios where the data is non-sensitive or the server can be trusted, this is less likely to be an issue. In other scenarios, techniques such as differential privacy (DP) \cite{dwork_DP_2014} or homomorphic encryption \cite{madi_homomorphic_2021} could be applied to reduce the risk of privacy breach. As pointed out in e.g. \cite{zhang_dense}, the fact that only a single communication round is needed reduces the risk of privacy leakage. Furthermore, the single-round setup would have an advantage in the context of DP. The entire privacy budget could be allocated to this single round of communication. This might mitigate the effects of depletion of privacy budget due to the number of communication rounds, a problem reported by e.g. the authors of \cite{huang_shuffled_dp_fl}.

In general, approaches to improving privacy in the context of EM algorithms include encryption \cite{leemaqz_ppdem_2017} , DP \cite{Diao_privgmm}, secure summation \cite{lin_securesummation_2005}, and subspace perturbations \cite{li_subspace_DEM_GMM}, which all add some complexity to the overall solution. Although several of these approaches could be applied to FedGenGMM to enhance privacy protection, we leave this as a future work.

\section{Experimental evaluation}

\subsection{Datasets}
We conducted the evaluation using one proprietary dataset ("VEHICLE"), in the form of vehicle time series data collected during the operation of Scania vehicles, and four public datasets from different domains, including image data (MNIST \cite{mnist_2012}), tabular data (Covertype \cite{covertype_1998}), and time series data (WADI \cite{wadi_2017}, RWHAR \cite{rwhar_2016}, and the Server Machine Dataset (SMD) \cite{omni_anomaly_smd_2019}). The basic characteristics of each dataset are outlined in table \ref{tab:datasets_overview}. Since high-dimensionality data can lead to high computational complexity in the GMM training process \cite{pfl_mixture}, we used principal component analysis (PCA) to reduce the number of features for MNIST (from 784 to 24) and RWHAR (from 63 to 16). In addition, to facilitate the creation of OOD data, we excluded all binary features from Covertype and from WADI, leaving 10 and 84 features for those datasets, respectively. For each dataset, the original number of features is given within brackets in the column "Features" in table \ref{tab:datasets_overview}. All features were normalized to the range of $[0, 1]$. 

\begin{table}
    \caption{Datasets employed for evaluation}
    \label{tab:datasets_overview}
    \centering
    \begin{tabular}{cccccc}
        \toprule
        Dataset & Domain & Size & Features & Classes & Partitioning\\
        \midrule
        MNIST & Handwritten digits & 60 000 & 24 (784) & 10 & Dir($\alpha$)\\
        Covertype & Tabular terrain data & 580 000 & 10 (54) & 7 & Dir($\alpha$)\\
        RWHAR & Human activity body sensors & 338 000 & 16 (63) & 13 & Dir($\alpha$)\\
        WADI & Water facility operations & 950 000 & 84 (123) & 10 & Quantity($\alpha$)\\
        VEHICLE & Air pressure system operations & 12 000 & 11 (11) & 3 & Quantity($\alpha$)\\
        SMD & Server machine operations & 1 400 000 & 38 (38) & 28 & Dir($\alpha$)\\
         \bottomrule
    \end{tabular}
\end{table}

\begin{table}
    \caption{Datasets and OOD settings}
    \label{tab:datasets_ood}
    \centering
    \begin{tabular}{ccccc}
        \toprule
        Dataset & Test set size & Inlier data & Anomalous data & Anomaly ratio\\
        \midrule
        MNIST & 12 000 & Original images & Manipulated images & 10\%\\
        Covertype & 116 000 & Original & Added Gaussian noise & 10\%\\
        RWHAR & 67 000 & Walking & Running & 10\%\\
        WADI & 170 000 & Normal ops & Cyber attack mode & 6\%\\
        VEHICLE & 3 000 & Normal ops & Induced air leakage & 50\%\\
        SMD & 700 000 & Normal ops & Observed malfunctions & 4\%\\ 
        \bottomrule
    \end{tabular}
\end{table}

\subsection{Data partitioning schemes}
\begin{figure}[h]
  \begin{subfigure}[t]{0.32\textwidth}
    \centering
    \includegraphics[width=\textwidth]{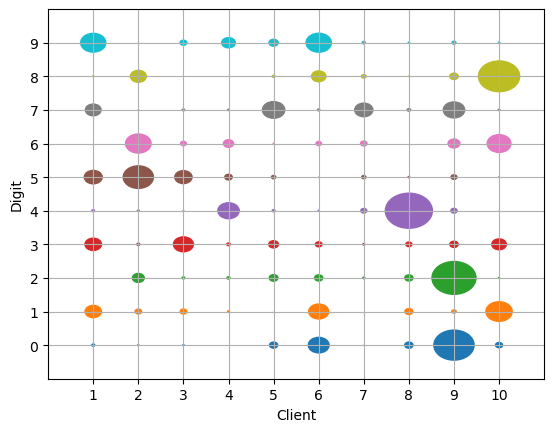}
    \caption{Dir($\alpha$=0.5)}
  \end{subfigure}
  \hfil
  \begin{subfigure}[t]{0.32\textwidth}
    \centering
    \includegraphics[width=\textwidth]{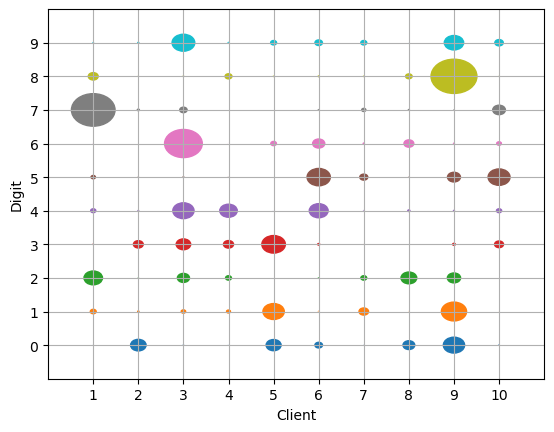}
    \caption{Dir($\alpha$=0.3)}
  \end{subfigure}
  \begin{subfigure}[t]{0.32\textwidth}
    \centering
    \includegraphics[width=\textwidth]{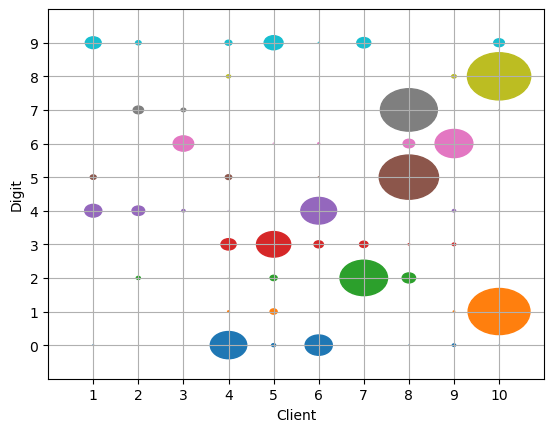}
    \caption{Dir($\alpha$=0.1)}
  \end{subfigure}
  \caption{Illustration of heterogeneity levels for different values of ($\alpha$), using Dir($\alpha$) to create partitions of the MNIST dataset for 10 clients. Each row of dots corresponds to one digit, i.e., one underlying distribution, and each column corresponds to one client. The size of the dot located at (x, y) is proportional to the number of datapoints from digit y seen by client x during training. With $\alpha=0.5$, most of the clients will use data from at least half of the digits. With $\alpha=0.1$, most of the data from each digit (i.e., digit 1 and 8) is assigned to only one or two clients. Note that the local dataset sizes also become more heterogeneous with decreasing $\alpha$.}
  \label{fig:client_partitions}
\end{figure}

To simulate heterogeneity (in the form of feature distribution skew) of the client distributions, we partitioned the data using two different schemes. In both schemes, each underlying distribution corresponds to one "class" in the dataset, e.g., digit "2" in the case of MNIST. The first scheme relies on drawing the mixture weights $\pi_{c,m}$ in Eq. \ref{eq:input_mixture} from a symmetric Dirichlet distribution with parameter $\alpha$, similar to other works such as \cite{Marfoq_2021} and \cite{zhang_dense}. Here, smaller values of $\alpha$ mean that the client distributions will be more heterogeneous. The effect of different values of $\alpha$ is illustrated in Figure \ref{fig:client_partitions} for the case of MNIST data distributed over 10 clients. In the following, we refer to this scheme as "Dir($\alpha$)". In the second scheme, each client mixture is created by assigning data from a fixed number (also denoted by $\alpha$) of randomly chosen underlying distributions $p^{(m)}$ to each client, similar to the heterogeneity scheme in \cite{modfl} and one of several schemes employed in \cite{Li2022_non_iid_silos} (termed "Quantity-based label imbalance"). We denote this scheme by "Quantity($\alpha$)".

As stated above, for MNIST, each class corresponds to data from images representing the same digit. For Covertype, we use the different terrain types, and for RWHAR each class corresponds to sensor data from a specific person. The SMD consists of data from 28 different servers, thus allowing for 28 classes in our experiments, and for the VEHICLE dataset, we create partitions based on which operational environment (i.e., city, high-way, or test track) the data was collected in. Finally, for the WADI experiments, the classes were created artificially. The data corresponding to the underlying distribution $p^{(m)}$ was created by adding values drawn from a multivariate Gaussian with center $C_m = \mathbf{1} (m-1) \beta$ and diagonal covariance 0.01 to the original data points, with $\beta$ being a parameter controlling the difference between the generated data for different classes. The same type of perturbation was applied to both the non-anomalous and anomalous parts of the dataset.

\subsection{Obtaining anomalous data}
Four of the datasets (RWHAR, WADI, VEHICLE, and SMD) were collected under conditions that allow a natural separation into inlier and anomalous (OOD) data subsets, as indicated in Table \ref{tab:datasets_ood}. For these cases, we used a subset of the inlier data as the training set to be partitioned and distributed to the clients. After training, the global model was applied to the rest of the inlier data together with a randomly chosen subset of the OOD data.

For the MNIST and Covertype datasets, we created artificial OOD data. Initially, the dataset was split into three parts, with the first part forming the training set, the second part the inlier part of the test set, and the third part making up the OOD part of the test set. The images in MNIST were rotated 90 degrees counter-clockwise, flipped horizontally, and scaled with a factor of 1.2. The Covertype terrain data was perturbed by adding Gaussian noise with zero mean and variance 0.005. For both datasets, the same OOD perturbation was applied across all classes.

\subsection{Baselines}
We compared the performance of FedGenGMM against local models, different versions of DEM for GMMs, and a non-federated model.

In the local model approach, the individual models trained on each local dataset $D_c$ were applied to the evaluation dataset and the resulting scores were averaged.

To obtain a stronger baseline, we also used a distributed version of the EM algorithm for GMM, as outlined in \cite{pfl_mixture}, to create the global model. Here, the number of components are assumed to be the same for the global model and all local models, and the training process starts by having the server choose initial GMM parameters and distribute them to the clients. Since the server does not have access to local data, it is not obvious how the initial GMM parameters should be chosen. We evaluated three different schemes for the initialization of the GMM component centers, (1) maximally separated centers given knowledge of the feature range \cite{Diao_privgmm}, (2) performing an initial GMM training round using a small subset of training data, and (3) using a federated version of k-means (from \cite{Dennis2021HeterogeneityFT}) for the estimation of the global component centers. Henceforth, we refer to these three variants as DEM init 1, DEM init 2, and DEM init 3, respectively. For DEM init 1, we used the fact that the features were normalized. For DEM init 2, the subset consisted of 100 data points drawn from the entire training dataset. We note that the second scheme involves sending data points to the server. The stopping criterion for the DEM iterations was based on the change in the average likelihood over the client models when applied to the local datasets.

Finally, as a benchmark representing the "ideal" solution, we used the EM algorithm to train the global model using the entire training dataset in a non-federated setup.

\subsection{Settings}
In order for the model sizes to be reasonable with respect to the size of the local datasets, and given the setting of edge computing, which implies limited computational resources in the clients, we chose to use diagonal covariance matrices for the GMMs (refer to Section \ref{sec:complexity} for complexity analysis).

For each dataset, we determined the number of GMM components in the global model ($K$) by training non-federated GMMs with varying $K$. We analyzed the average log-likelihood resulting from applying the trained model to a validation part of the dataset, and a value representing a good trade-off between model size and performance was chosen. The resulting values are given in Table \ref{tab:datasets_settings}. Furthermore, for most of the experiments, we let $K_c = K$ for all FedGenGMM clients to facilitate comparisons between different federated algorithms. We note that this way of determining $K_c$ would not be feasible in a federated setting, where no single party would have access to a representative part of the dataset. A more realistic approach would be to let each client $c$ estimate $K_c$ using its local data, something that could easily be accommodated in FedGenGMM since the method allows heterogeneous local models.

Unless otherwise stated, the number of participating clients was set to 20 for all datasets except VEHICLE, where we used 12 clients due to the limited dataset size.
\begin{table}
    \caption{Datasets and settings}
    \label{tab:datasets_settings}
    \centering
    \begin{tabular}{cccccc}
        \toprule
        Dataset & $K$ & Clients\\
        \midrule
        MNIST & 30 & 20\\
        Covertype & 15 & 20\\
        RWHAR & 15 & 20\\
        WADI & 10 & 20\\
        VEHICLE & 15 & 12\\
        SMD & 10 & 20\\
        \bottomrule
    \end{tabular}
\end{table}

The initialization of the local GMM components was done using k-means on local data.  For the generation of synthetic data in the server, we set the value of $H$ to 100 in Eq. \ref{eq:ns}. The convergence limit (in terms of the difference in the likelihood of the training data between two consecutive EM iterations) for the GMM training was set at $10^{-3}$ throughout the experiments, both for FedGenGMM, the DEM algorithms and the non-federated benchmark.

Each run (consisting of dataset generation, federated training, and evaluation) was repeated five times to obtain the mean and variance of the performance scores for the different methods and scenarios. In each run, a new split of the dataset into train and test subsets was created, as well as a new partition of the data between clients and new OOD data for the anomaly detection experiments.

\subsection{Results - fitting a model to the global distribution}
\begin{figure}[h]
  \begin{subfigure}[t]{0.45\textwidth}
    \centering
    \includegraphics[width=\textwidth]{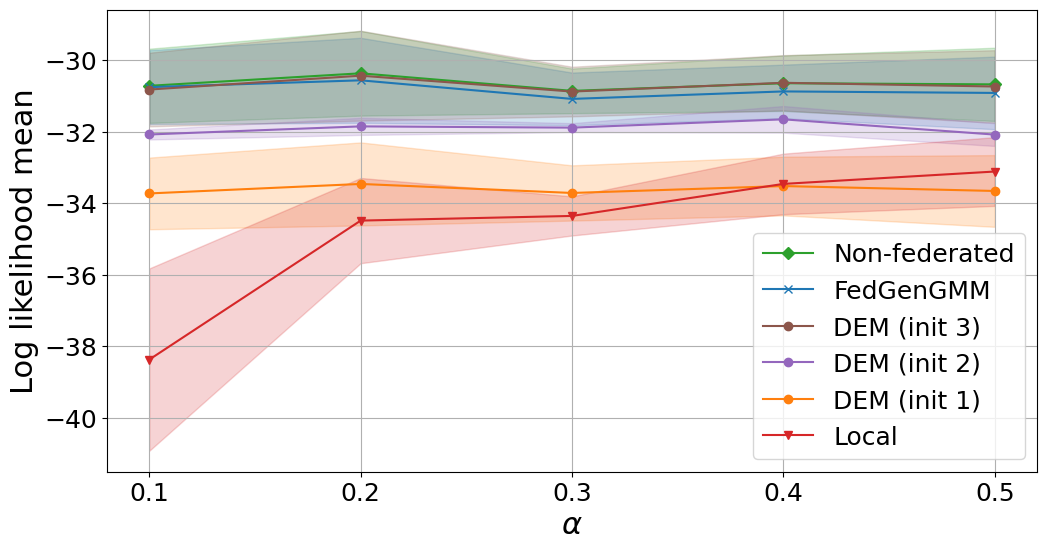}
    \caption{MNIST}
    \label{fig:ll-mnist}
  \end{subfigure}
  \hfil
  \begin{subfigure}[t]{0.45\textwidth}
    \centering
    \includegraphics[width=\textwidth]{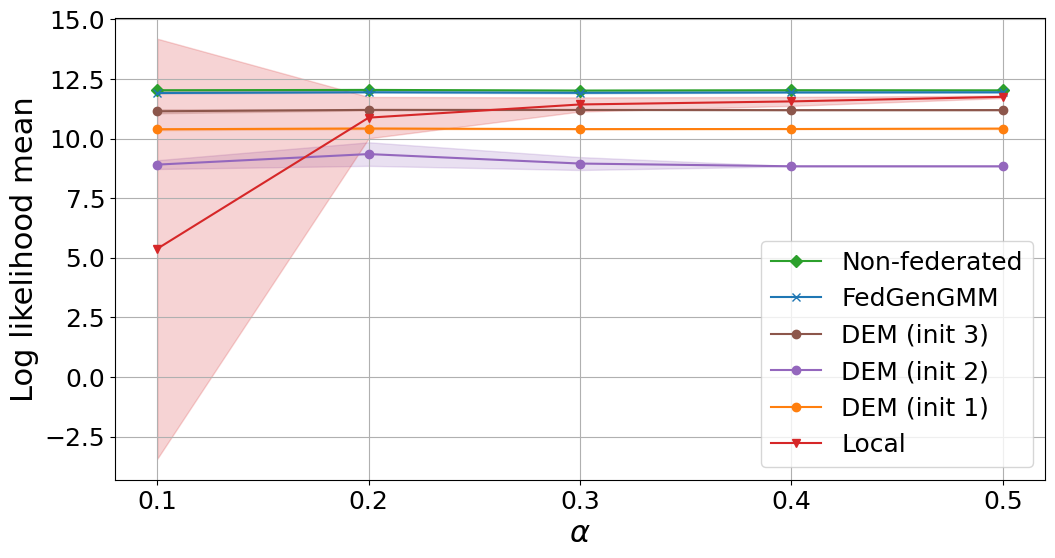}
    \caption{Covertype}
    \label{fig:ll-covertype}
  \end{subfigure}
  \hfil
  \begin{subfigure}[t]{0.45\textwidth}
    \centering
    \includegraphics[width=\textwidth]{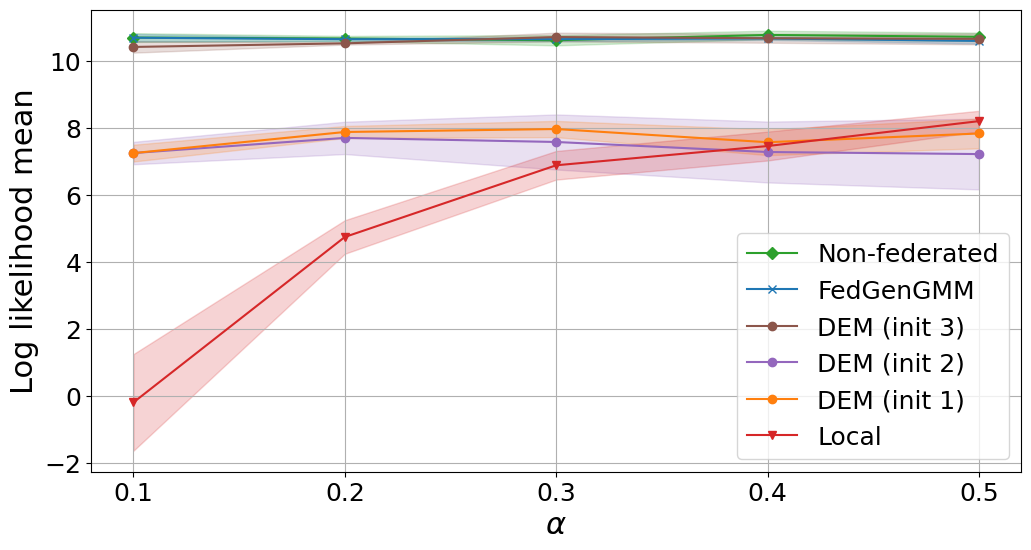}
    \caption{RWHAR}
    \label{fig:ll-rwhar}
  \end{subfigure}
  \begin{subfigure}[t]{0.45\textwidth}
    \centering
    \includegraphics[width=\textwidth]{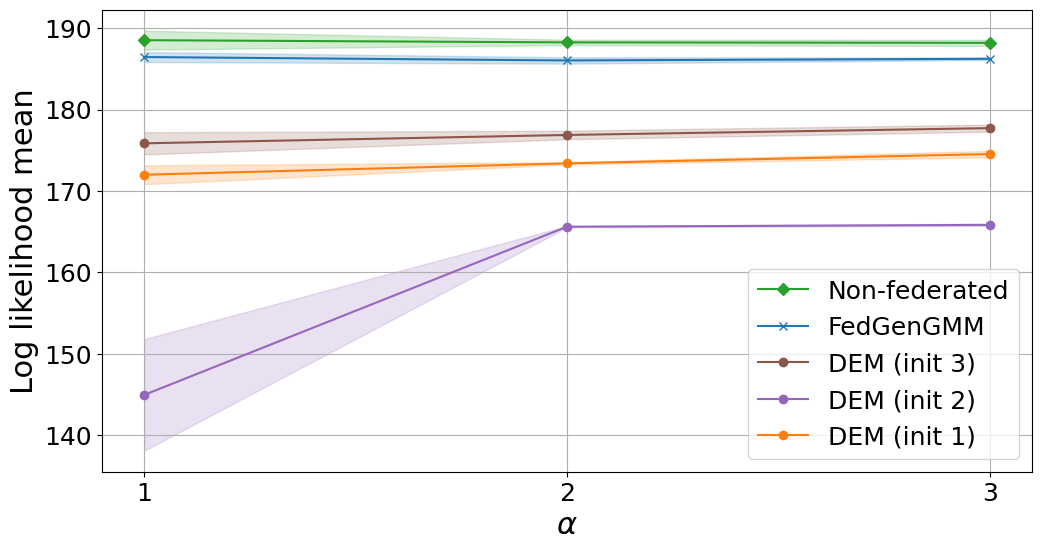}
    \caption{WADI}
    \label{fig:ll-wadi}
  \end{subfigure}
  \hfil
  \begin{subfigure}[t]{0.45\textwidth}
    \centering
    \includegraphics[width=\textwidth]{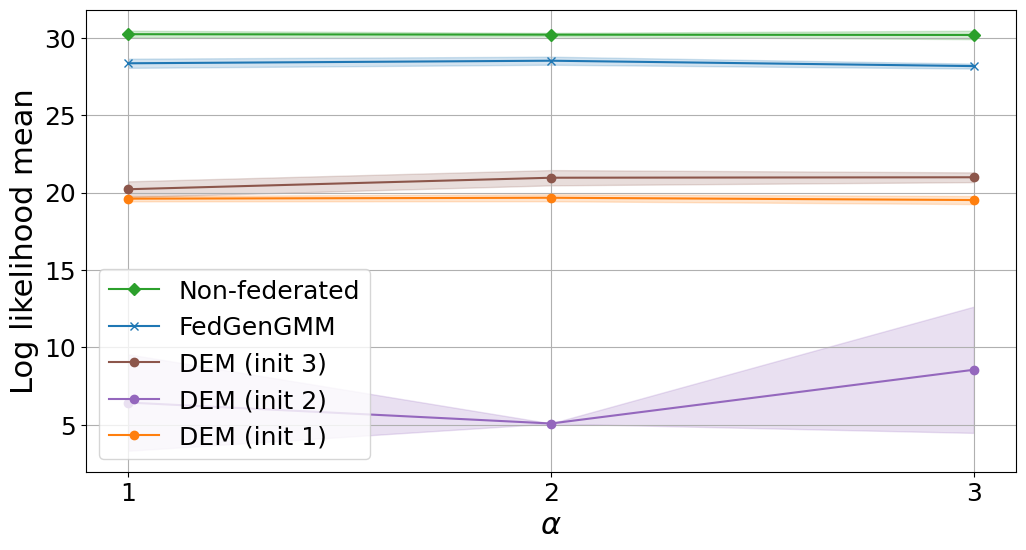}
    \caption{VEHICLE}
    \label{fig:ll-vehicle}
  \end{subfigure}
  \hfil
  \begin{subfigure}[t]{0.45\textwidth}
    \centering
    \includegraphics[width=\textwidth]{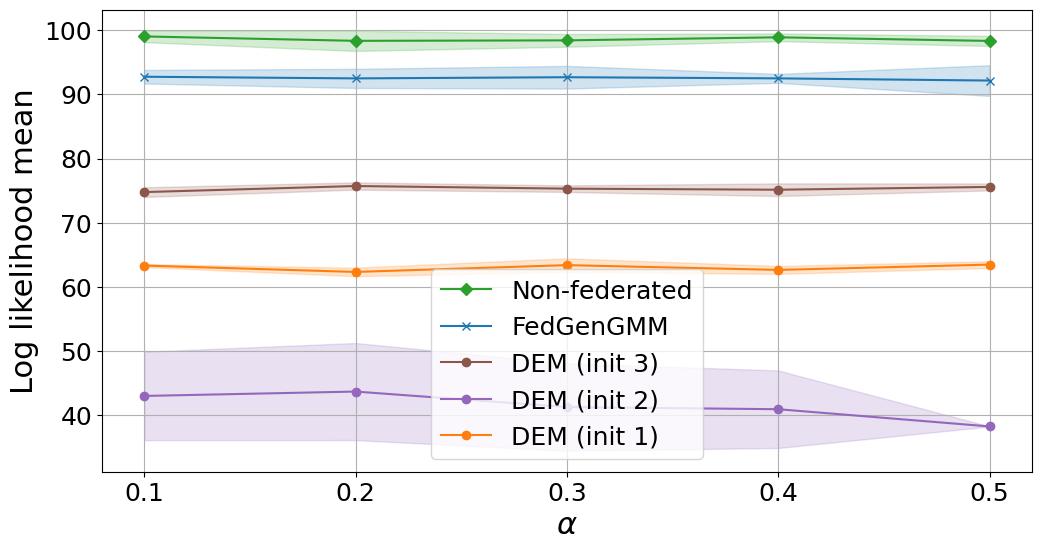}
    \caption{SMD}
    \label{fig:ll-smd}
  \end{subfigure}
  \caption{Resulting global model log likelihood mean for each dataset and method, using fixed $K$ with varying heterogeneity level ($\alpha$) of client data distributions. For WADI and VEHICLE, the heterogeneity scheme is Quantity($\alpha$), for others it is Dir($\alpha$). The markers indicate the mean over five runs, and the height of the shaded areas correspond to +/- the standard deviation. Results for local models are excluded for plots (d) - (f) due to the low values of the log likelihood mean.}
  \label{fig:likelihoods}
\end{figure}

The purpose of the first set of experiments was to evaluate the ability of FedGenGMM to produce a global model that captures the patterns of the global distribution, under varying heterogeneous conditions, and for each of the six datasets included in the study. As performance score, we used the average log-likelihood produced by applying the resulting model to the entire training dataset. For each dataset, a fixed number of GMM components ($K$) was used, as indicated in table \ref{tab:datasets_settings}. The results are shown in Figure \ref{fig:likelihoods}.

As can be seen from the plots, all the methods except the local approach have stable performance under the different degrees of heterogeneity included in the experiments. Among the three different versions of DEM, init 3 (which relies on distributed k-means for the initialization of the GMM component centers) performs best for all datasets. The ability of FedGenGMM to adapt to the global distribution is persistently on par with or better than the best of the DEM methods and is in fact close to the benchmark for all datasets. In terms of result variance, both FedGenGMM and DEM init 3 exhibit relatively low variance for all cases except MNIST, possibly due to the more complicated patterns present in that dataset, coupled with the relatively small size of the dataset. As expected, the local model approach struggles under conditions of strong heterogeneity. The complete plots for (d) - (f), including the local model performance, can be found in Appendix \ref{appendix:A}.

\subsection{Comparison of FedGenGMM with DEM}

\subsubsection{Communication cost}
For comparison of communication cost, the number of iterations required for the different DEM versions was recorded. As shown in Table \ref{tab:dem_iterations}, this number varies both with the dataset and with the type of parameter initialization. Overall, the number of communication rounds required for DEM is around one order of magnitude higher than that of the one-shot FedGenGMM process. The actual number of iterations required for DEM in a specific case would depend on several factors, such as the nature of the dataset, the initial parameters of the GMMs, and the tolerance limits set for convergence. However, in general (for centralized EM), the number of iterations can be said to be between 10 and 100 \cite{figueiredo_2002_mixture_models}.

\begin{table}[h]
    \caption{Average number of communication rounds required for training the global model, recorded over five runs for each combination of dataset, algorithm and heterogeneity level (heterogeneity levels as in Figure \ref{fig:likelihoods}).}
    \label{tab:dem_iterations}
    \centering
    \begin{tabular}{cccccc}
        \toprule
        Dataset & FedGenGMM & DEM init 1 & DEM init 2 & DEM init 3\\
        \midrule
        MNIST & 1 & 33 & 41 & 38\\
        Covertype & 1 & 21 & 7 & 40\\
        RWHAR & 1 & 20 & 27 & 15\\
        WADI & 1 & 25 & 4 & 29\\
        VEHICLE & 1 & 29 & 3 & 18\\
        SMD & 1 & 36 & 12 & 30\\
        \bottomrule
    \end{tabular}
\end{table}

\subsubsection{Computational complexity}
The DEM algorithm performs mainly the same type of calculations as the centralized EM algorithm, with the E-step and M-step being executed locally in each client and the resulting parameters being aggregated in the server after each iteration. Compared to FedGenGMM, DEM thus adds the server aggregation part at the end of each iteration. On the other hand, FedGenGMM requires the generation of synthetic data and one full execution of the standard EM-algorithm on the server. Thus, while server-side complexity can be expected to be higher for FedGenGMM, computational complexity in the clients (which is most relevant in the context of edge computing) should be similar for the two approaches.

\subsubsection{Privacy}
In FedGenGMM, the local models are trained until convergence and the resulting parameters are sent to the server, something that may raise privacy concerns, as discussed in Section \ref{sec:privacy}. The process is a bit different from the DEM algorithms, where after each local EM iteration, the parameters are sent to the server. In the server, they are aggregated over all participating clients. Then, the aggregated parameters are sent back to each client for the next iteration, and the process is repeated until convergence is reached. However, using DEM algorithms might still enable the server to reconstruct local statistics to a large extent, as shown in \cite{li_subspace_DEM_GMM}.

\subsection{Results - anomaly detection}
In the anomaly detection scenario, the model to be evaluated was used to produce point-wise log-likelihood values for a hold-out dataset. The hold-out dataset consisted of one part ID data and one part OOD data, with the ratios given in Table \ref{tab:datasets_ood}. The resulting AUC-PR was used as performance score. The main purpose of the experiments was to show that for a given GMM configuration (that is, given $K$ and the type of covariance matrix), the one-shot FedGenGMM method can achieve performance at least similar to DEM methods under heterogeneous conditions. However, we also investigated the possibility of using smaller client models during the FedGenGMM learning phase and then aggregating them into a larger global model. Furthermore, the impact of the number of clients was briefly studied.

Although several of the datasets consist of time series that exhibit both point-wise and sequential anomalies, in the present study all anomaly detection scores were produced as point-wise scores. A combination of point-wise scoring with windowing techniques, such as smoothing the input and using moving averages of anomaly scores, would likely increase the performance. However, this would also introduce more hyperparameters in the experiments. Given the main purpose of analyzing the relative performance of the methods, no more elaborate schemes were implemented. 

\begin{figure}[h]
  \begin{subfigure}[t]{0.45\textwidth}
    \centering
    \includegraphics[width=\textwidth]{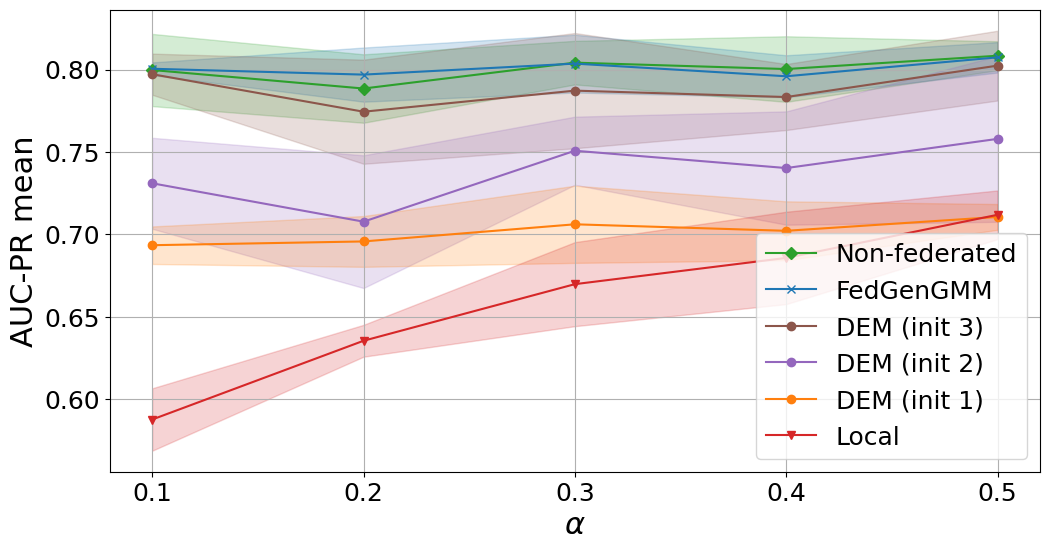}
    \caption{MNIST}
    \label{fig:pr-mnist}
  \end{subfigure}
  \hfil
  \begin{subfigure}[t]{0.45\textwidth}
    \centering
    \includegraphics[width=\textwidth]{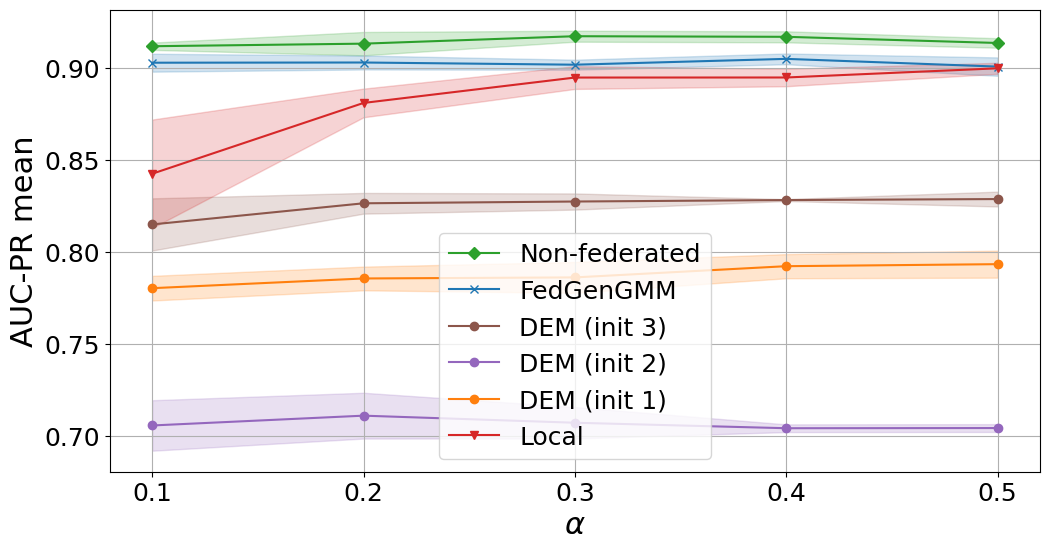}
    \caption{Covertype}
    \label{fig:pr-covertype}
  \end{subfigure}
  \hfil
  \begin{subfigure}[t]{0.45\textwidth}
    \centering
    \hspace*{-0.2in}
    \includegraphics[width=\textwidth]{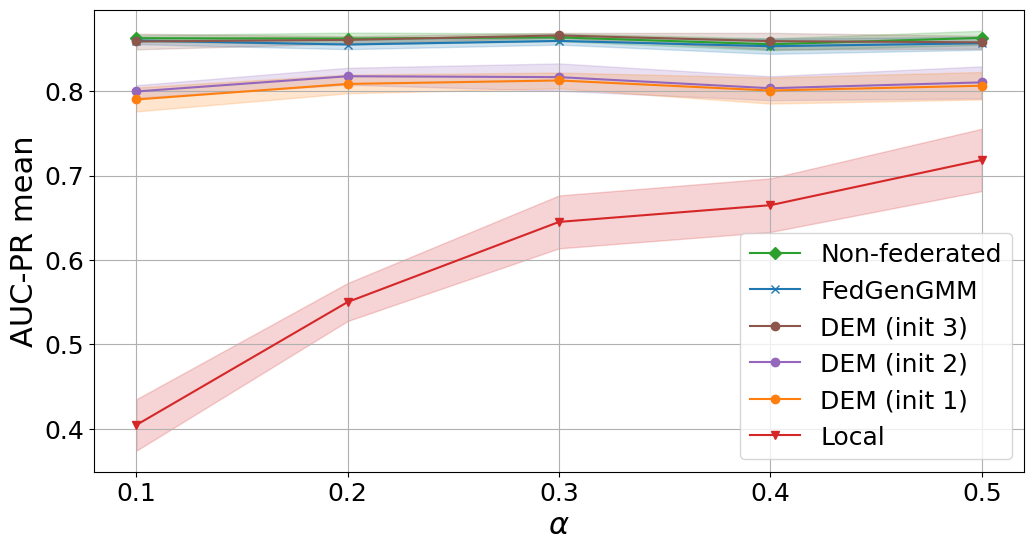}
    \caption{RWHAR}
    \label{fig:pr-rwhar}
  \end{subfigure}
  \begin{subfigure}[t]{0.45\textwidth}
    \centering
    \includegraphics[width=\textwidth]{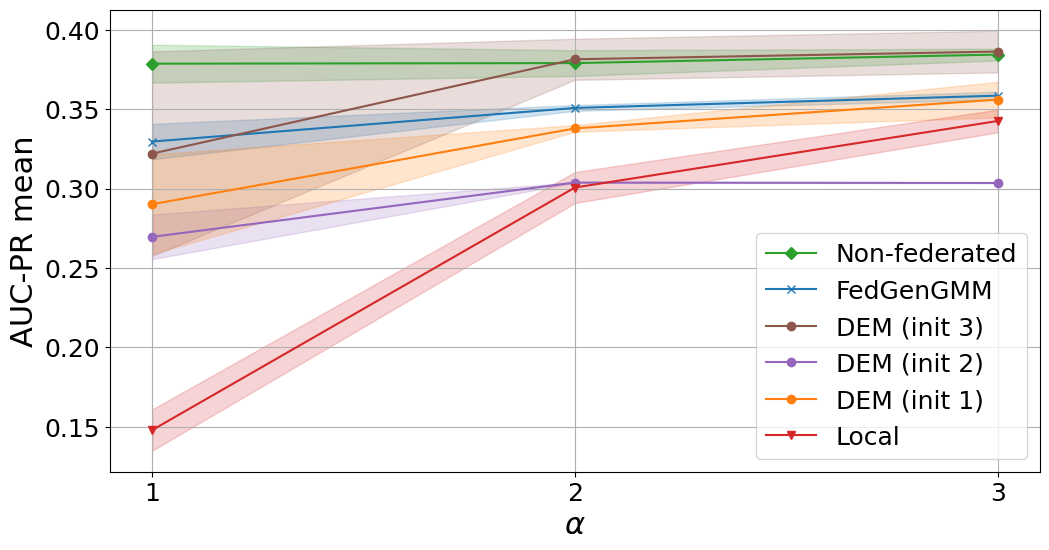}
    \caption{WADI}
    \label{fig:pr-wadi}
  \end{subfigure}
  \hfil
  \begin{subfigure}[t]{0.45\textwidth}
    \centering
    \includegraphics[width=\textwidth]{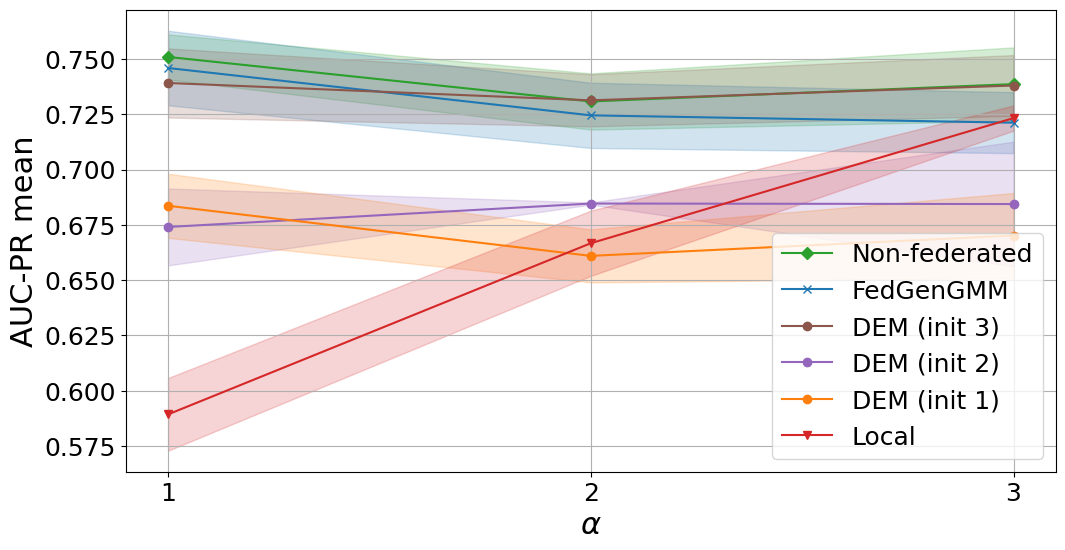}
    \caption{VEHICLE}
    \label{fig:pr-vehicle}
  \end{subfigure}
  \hfil
  \begin{subfigure}[t]{0.45\textwidth}
    \centering
    \includegraphics[width=\textwidth]{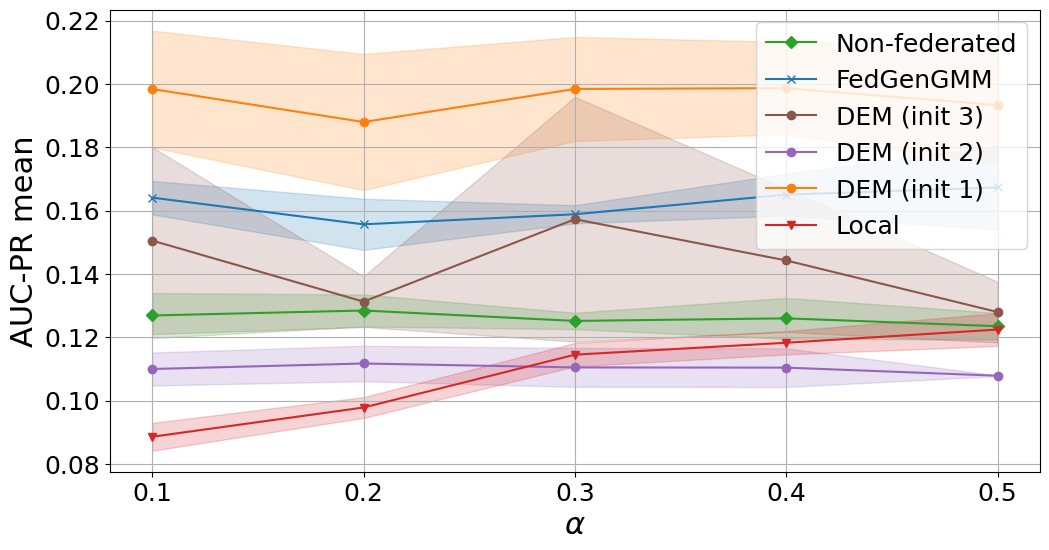}
    \caption{SMD}
    \label{fig:pr-smd}
  \end{subfigure}
  \caption{Anomaly detection AUC-PR mean for each dataset and method, using fixed $K$ with varying heterogeneity level ($\alpha$) of client data distributions. For WADI and VEHICLE, the heterogeneity scheme is Quantity($\alpha$), for others it is Dir($\alpha$). The markers indicate the mean over five runs, and the height of the shaded areas correspond to +/- the standard deviation.}
  \label{fig:auc-pr-alpha}
\end{figure}

\textbf{Varying heterogeneity}
In this set of experiments, we evaluated the ability of the methods to detect anomalous data points, while varying the heterogeneity of the client data distributions. The results presented in Figure \ref{fig:auc-pr-alpha} indicate that the performance of the three DEM methods, as well as that of FedGenGMM, is relatively insensitive to the level of heterogeneity.

FedGenGMM performs at least on par with the DEM methods in all cases except for SMD, outperforming them for Covertype, where all three DEM variants struggle. The results for SMD stand out somewhat, with three of the methods actually achieving higher average scores than the benchmark, which uses nonfederated training. In most of the cases (especially for SMD), FedGenGMM exhibits a consistently lower variance than the DEM methods.

\begin{figure}[h]
  \begin{subfigure}[t]{0.48\textwidth}
    \centering
    \includegraphics[width=\textwidth]{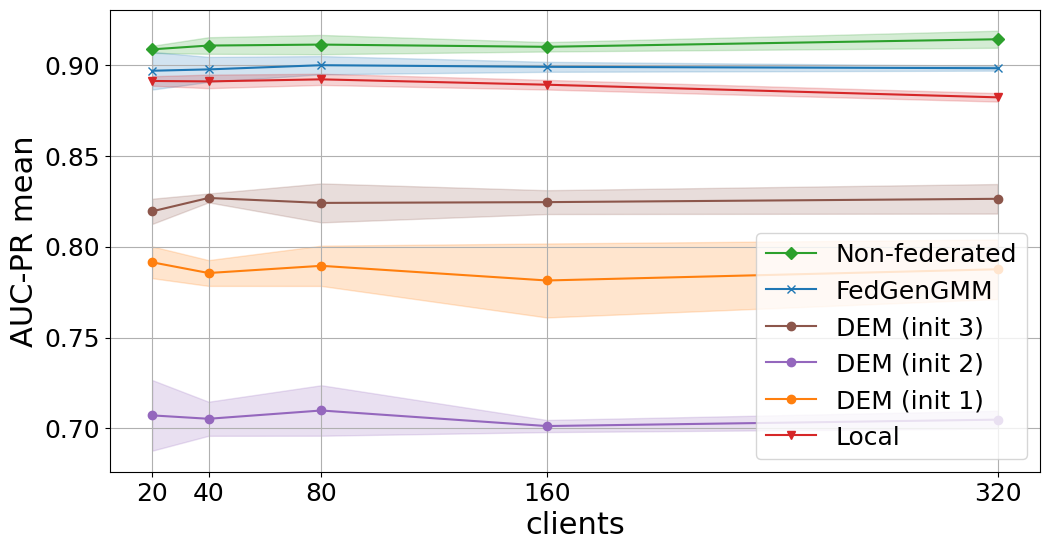}
    \caption{Covertype}
    \label{fig:pr-nc-covertype}
  \end{subfigure}
  \hfil
  \begin{subfigure}[t]{0.48\textwidth}
    \centering
    \includegraphics[width=\textwidth]{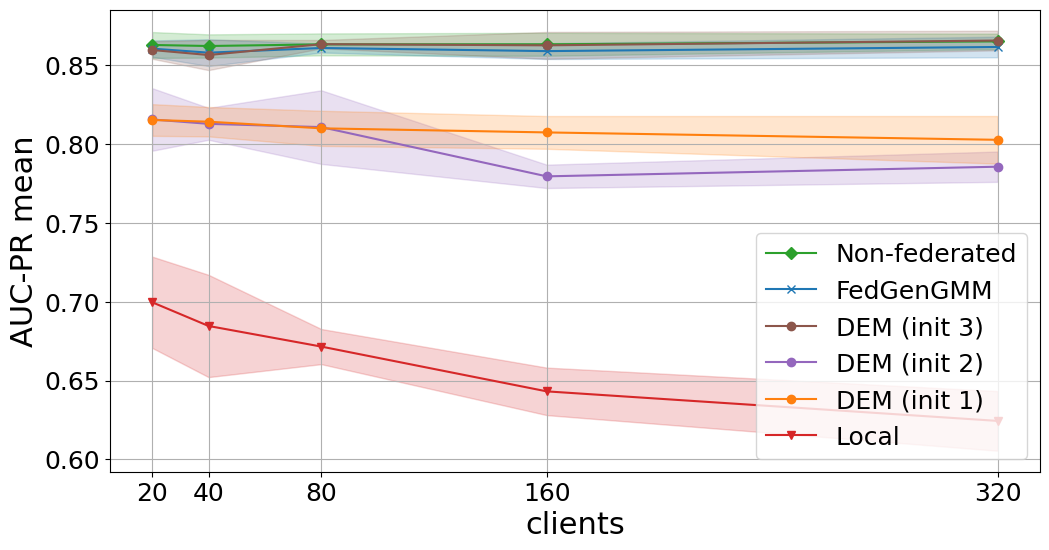}
    \caption{RWHAR}
    \label{fig:pr-nc-rwhar}
  \end{subfigure}
  \begin{subfigure}[t]{0.48\textwidth}
    \centering
    \includegraphics[width=\textwidth]{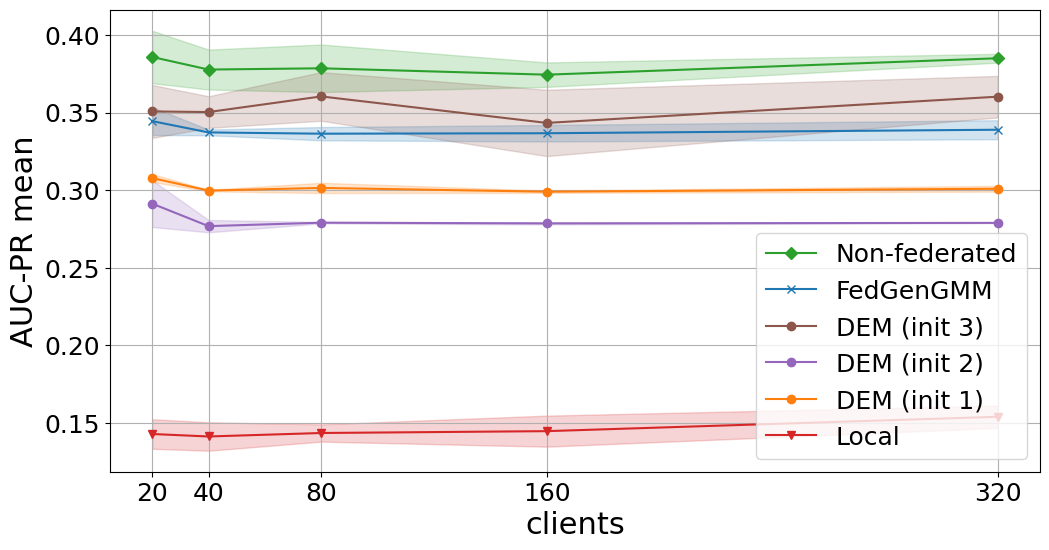}
    \caption{WADI}
    \label{fig:pr-nc-wadi}
  \end{subfigure}
  \hfil
  \begin{subfigure}[t]{0.48\textwidth}
    \centering
    \includegraphics[width=\textwidth]{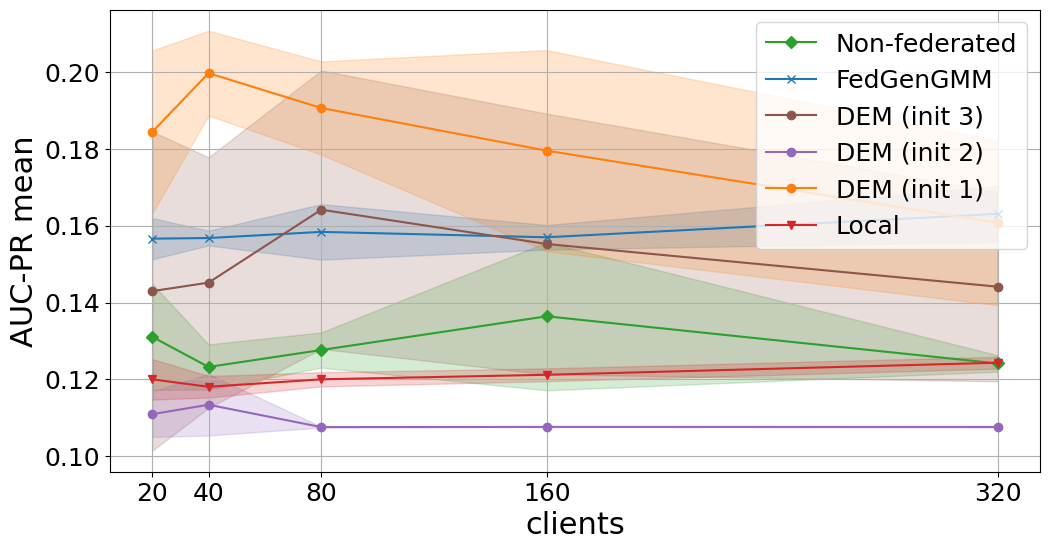}
    \caption{SMD}
    \label{fig:pr-nc-smd}
  \end{subfigure}
  \caption{Anomaly detection AUC-PR mean for each dataset and method, varying the number of clients from 20 to 320 while using fixed $K$ and fixed heterogeneity levels. For WADI, the heterogeneity scheme is Quantity($\alpha$=1), for others it is Dir($\alpha$=0.2). The markers indicate the mean over five runs, and the height of the shaded areas correspond to +/- the standard deviation. The datasets MNIST and VEHICLE were not included here due to their relatively small sizes, providing too few datapoints per client for larger number of clients.}
  \label{fig:auc-pr-n_clients}
\end{figure}

\textbf{Varying number of clients}
The impact of an increasing number of clients was evaluated in a separate experiment, where the settings for heterogeneity and GMM configurations were kept constant for each dataset. The results are illustrated in Figure \ref{fig:auc-pr-n_clients}. Overall, all the methods show stable performance for the range of client set sizes included in the experiment.

\textbf{Constrained client models}
Given the context of resource-constrained clients or devices, a possible use case for FedGenGMM would be the use of less complex local models during the learning process while aggregating the local models into a larger global one. The clients would still have to use the larger model for inference, but the resource requirements for GMM inference are typically much lower than for GMM training when using the EM-algorithm, since the training entails many iterations. One example is that since the computational complexity in the clients during training is linear in the number of GMM components, using fewer components locally would reduce the training complexity. In the experiment, the number of GMM components for the local models was varied from 2 to 20 while the global model consistently comprised 20 components (both the federated model and the non-federated benchmark). For comparison, we also show the performance of the DEM init 3 (k-means initialization) method, with both local and global models restricted in the number of components.   

\begin{figure}[h]
  \begin{subfigure}[t]{0.45\textwidth}
    \centering
    \includegraphics[width=\textwidth]{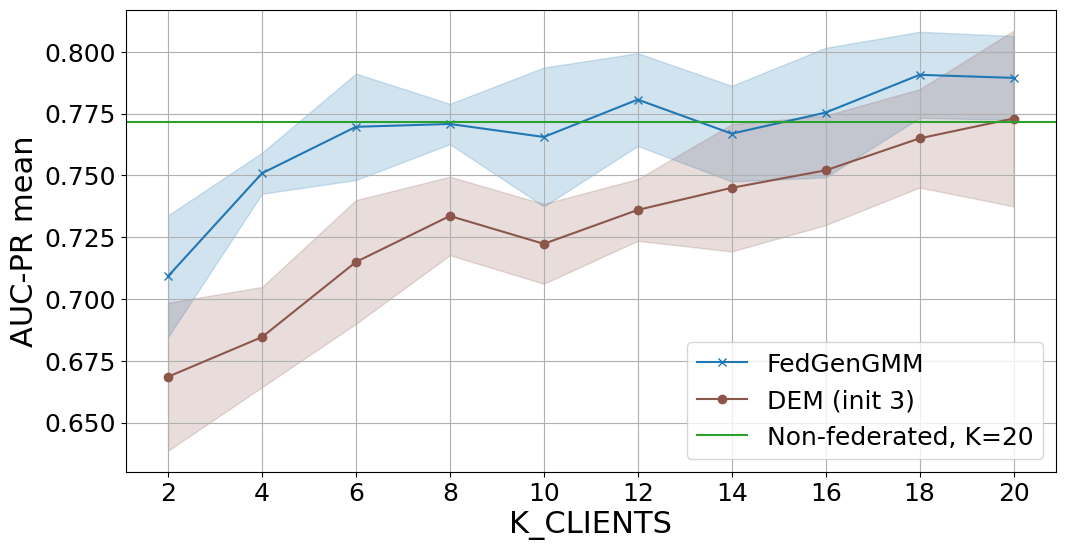}
    \caption{MNIST}
    \label{fig:pr-constr-mnist}
  \end{subfigure}
  \hfil
  \begin{subfigure}[t]{0.45\textwidth}
    \centering
    \includegraphics[width=\textwidth]{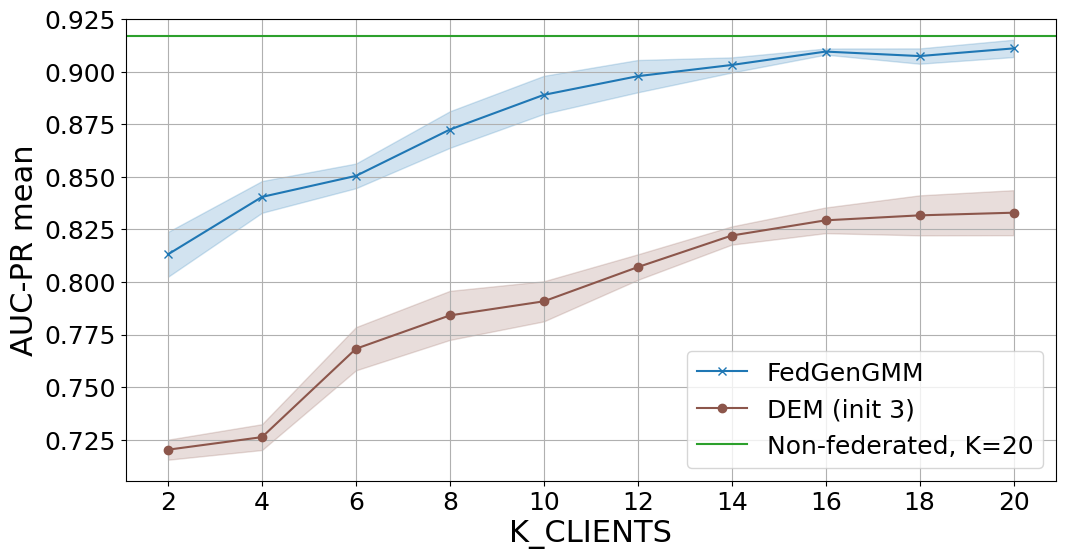}
    \caption{Covertype}
    \label{fig:pr-constr-covertype}
  \end{subfigure}
  \hfil
  \begin{subfigure}[t]{0.45\textwidth}
    \centering
    \includegraphics[width=\textwidth]{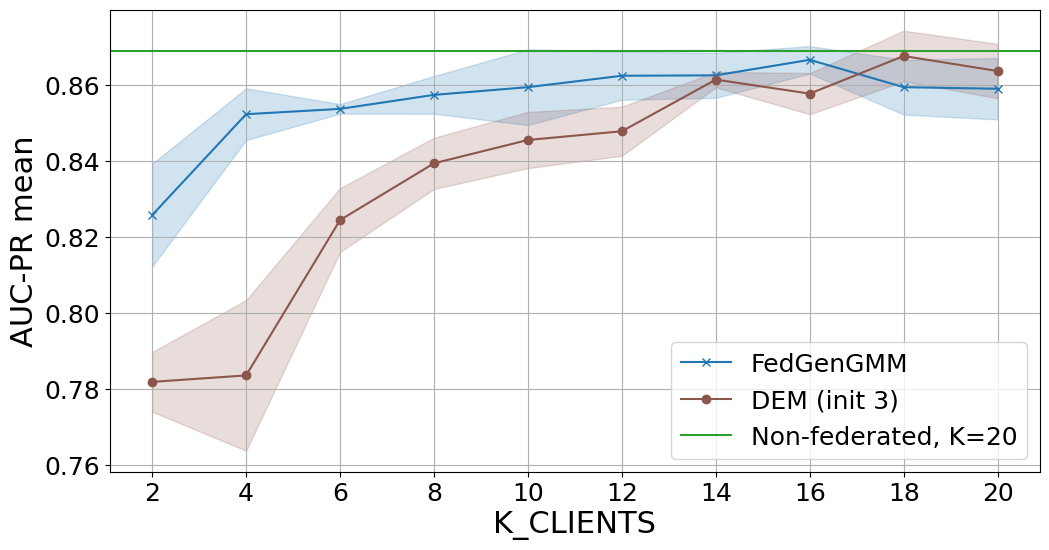}
    \caption{RWHAR}
    \label{fig:pr-constr-rwhar}
  \end{subfigure}
  \begin{subfigure}[t]{0.45\textwidth}
    \centering
    \includegraphics[width=\textwidth]{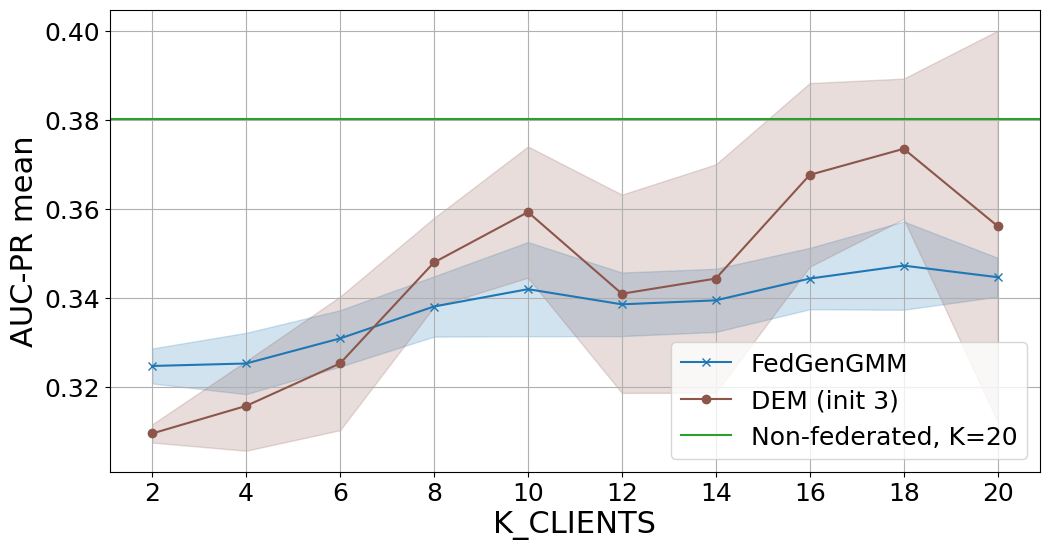}
    \caption{WADI}
    \label{fig:pr-constr-wadi}
  \end{subfigure}
  \hfil
  \begin{subfigure}[t]{0.45\textwidth}
    \centering
    \includegraphics[width=\textwidth]{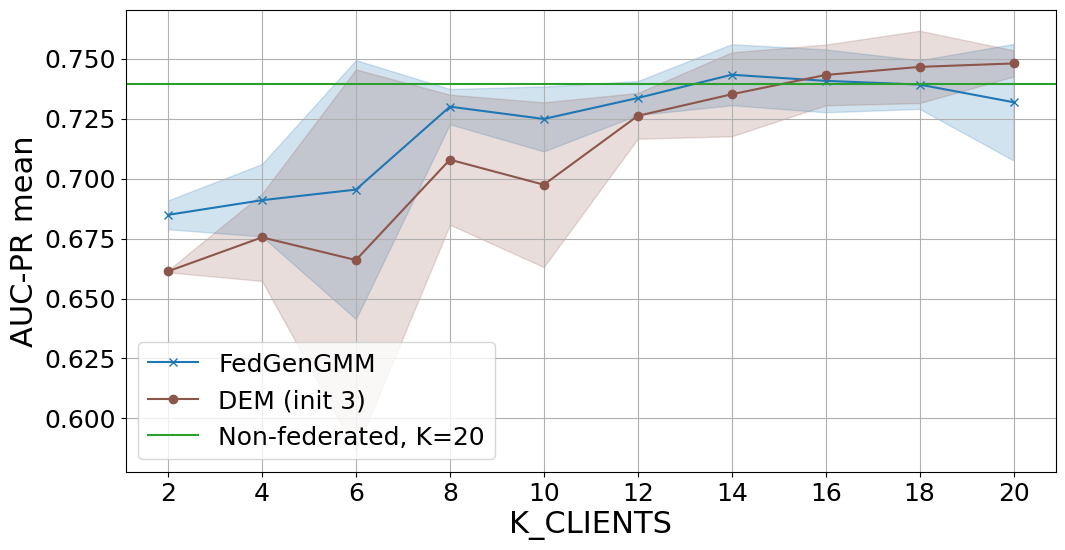}
    \caption{VEHICLE}
    \label{fig:pr-constr-vehicle}
  \end{subfigure}
  \hfil
  \begin{subfigure}[t]{0.45\textwidth}
    \centering
    \includegraphics[width=\textwidth]{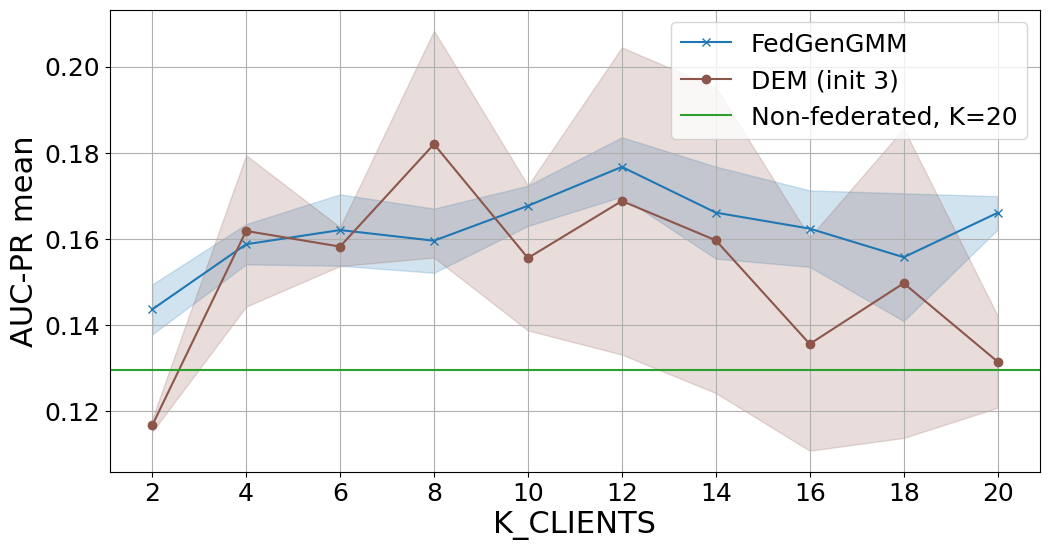}
    \caption{SMD}
    \label{fig:pr-constr-smd}
  \end{subfigure}
  \caption{Anomaly detection AUC-PR mean for each dataset and method, when varying the number of GMM components ($\text{K\_CLIENTS}$) in the local models. The global model for FedGenGMM uses fixed $K=20$, whereas the global DEM model uses $K=\text{K\_CLIENTS}$. For WADI and VEHICLE, the  heterogeneity scheme is Quantity($\alpha = 1$), for others it is Dir($\alpha = 0.2$). The markers indicate the mean over five runs, and the height of the shaded areas correspond to +/- the standard deviation. The green line indicates the average performance of the non-federated benchmark, using fixed $K=20$.}
  \label{fig:auc-pr-constr-k-clients}
\end{figure}

The results are illustrated in Figure \ref{fig:auc-pr-constr-k-clients}. In cases (a), (b), (c), and (e), FedGenGMM with constrained client models aggregated into a large central model achieves similar performance as the non-federated benchmark with full number of components (20). For MNIST (\ref{fig:pr-constr-mnist}) and Covertype (\ref{fig:pr-constr-covertype}), using approximately half the number of components locally achieves the same performance as the benchmark. However, the results are inconclusive for WADI and SMD, where the performance is largely unrelated to $K$. In most cases, FedGenGMM outperforms the DEM method for any given level of computational complexity of client training (that is, for any given number of local GMM components). However, when comparing the performance of FedGenGMM and DEM in this experiment, it should be noted that the DEM method uses a smaller global model than FedGenGMM.

\section{Conclusions}
In this work, we present and evaluate FedGenGMM, a method for one-shot federated training of Gaussian mixture models. The method uses the standard EM algorithm for client-side training and employs the generative property of the GMM to aggregate local models into a shared global model. We evaluate the method under controlled heterogeneous conditions, using six datasets of varying nature. Compared to distributed versions of the EM algorithm, FedGenGMM is shown to incur significantly lower communication costs. The results further indicate that the ability of FedGenGMM to learn the global data distribution in a federated setting is on par with that of a non-federated approach and better than the distributed EM methods included in the study. We also apply FedGenGMM to the problem of unsupervised FL for anomaly detection, with results showing performance being close to that of a non-federated model, at least on par with the distributed EM methods, and stable with respect to heterogeneity level and the number of clients. The flexibility of FedGenGMM in terms of allowing for different client model structures is further shown to have potential in reducing the computational complexity without sacrificing performance, compared to the less flexible approach of distributed EM. More work is required to consider potential privacy attacks and the mitigation of such attacks, to assess the potential of integrating other versions of the EM algorithm for local and central training, and to investigate the feasibility of applying the FedGenGMM concept to the problem of continuous federated learning.

\section*{Acknowledgements}

We thank Erik Pettersson for insights and useful discussions. This work was supported by Vinnova within the FFI program under contract 2020-02916.

\bibliographystyle{ACM-Reference-Format}
\bibliography{references}


\newpage
\appendix

\section{Full plots for estimation of the global distribution}
\label{appendix:A}
These plots include the log-likelihood mean of the local model approach (which was excluded from the plots in the main text). 
\begin{figure}[h]
  \begin{subfigure}[t]{0.5\textwidth}
    \centering
    \includegraphics[width=\textwidth]{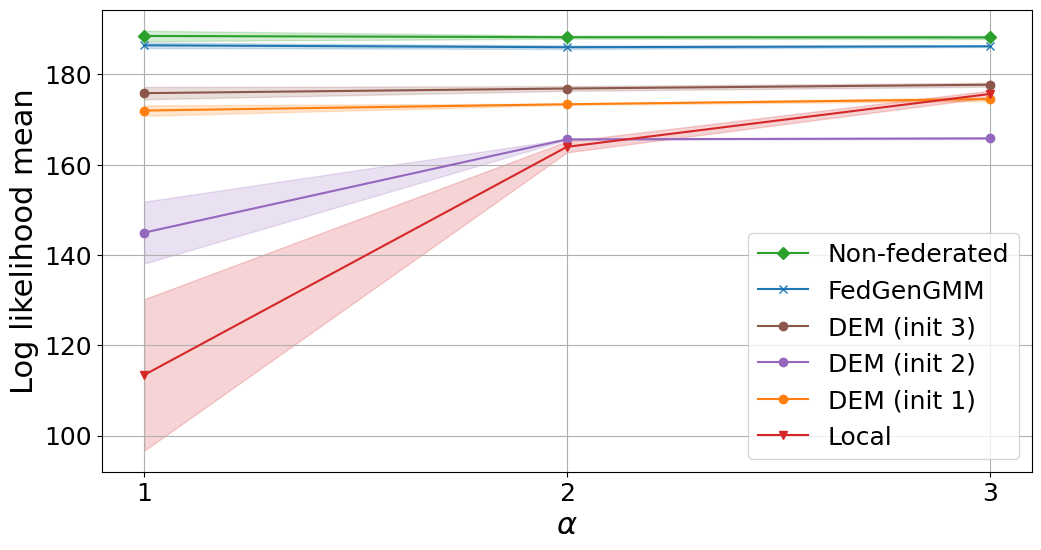}
    \caption{WADI}
    \label{fig:ll-wadi-full}
  \end{subfigure}
  \begin{subfigure}[t]{0.5\textwidth}
    \centering
    \includegraphics[width=\textwidth]{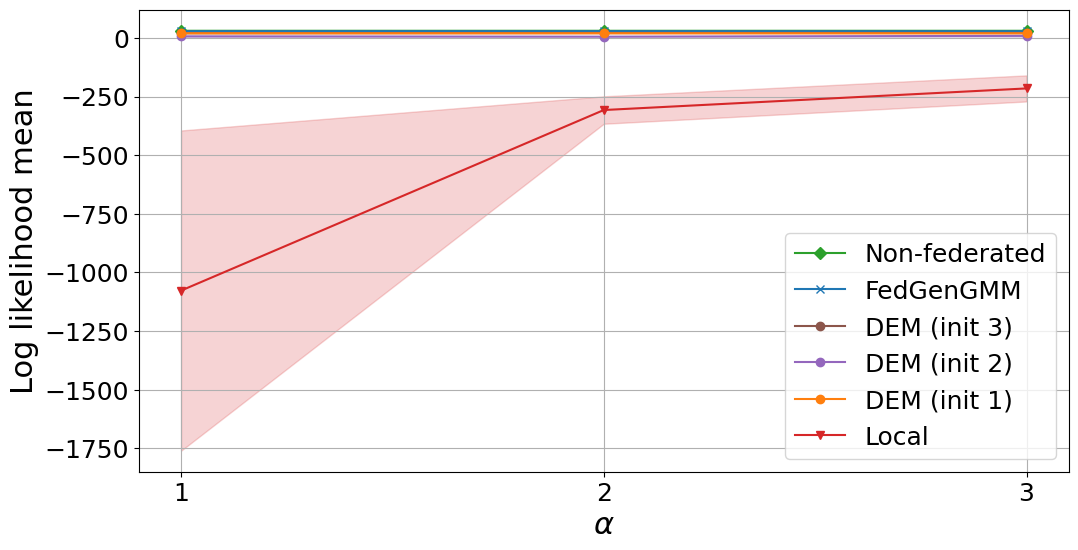}
    \caption{VEHICLE}
    \label{fig:ll-coda-full}
  \end{subfigure}
  \begin{subfigure}[t]{0.5\textwidth}
    \centering
    \includegraphics[width=\textwidth]{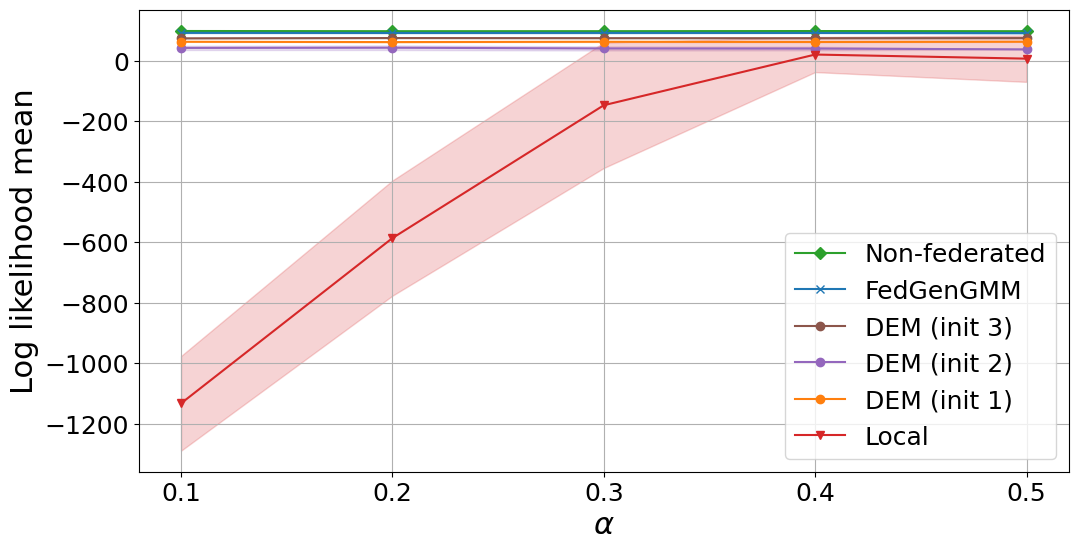}
    \caption{SMD}
    \label{fig:ll-smd-full}
  \end{subfigure}
  \caption{Resulting global model log likelihood mean for each dataset and method, using fixed $K$ with varying heterogeneity level (alpha) of client data distributions. For WADI and VEHICLE, the heterogeneity scheme is Quantity(alpha), for SMD it is Dir(alpha). The markers indicate the mean over five runs, and the height of the shaded areas correspond to +/- the standard deviation.}
  \label{fig:likelihoods_full}
\end{figure}

\end{document}